\documentclass{article}
\usepackage{PRIMEarxiv}
\usepackage[utf8]{inputenc} 
\usepackage[T1]{fontenc}    
\usepackage{hyperref}       
\usepackage{url}            
\usepackage{booktabs}       
\usepackage{amsfonts}       
\usepackage{nicefrac}       
\usepackage{microtype}      
\usepackage{lipsum}
\usepackage{fancyhdr}       
\usepackage{graphicx}       
\graphicspath{{media/}}     
\usepackage{amsmath} 

\title{Evaluating Neural Networks for Early Maritime Threat Detection}

\author{
  Dhanush Tella\thanks{Corresponding author: telladh@gmail.com}, Chandra Teja Tiriveedhi, Naphtali Rishe, \\ Dan E.\ Tamir, Jonathan I.\ Tamir \\
  Lake Highland Preparatory School, Orlando, FL \\
  University of Central Florida, Orlando, FL \\
  Florida International University, Miami, FL \\
  Texas State University, San Marcos, TX \\
  The University of Texas at Austin, Austin, TX
}

\begin{document}
\maketitle

\begin{abstract}
\textbf{\textit{Abstract}:} We consider the task of classifying trajectories of boat activities as a proxy for assessing maritime threats. Previous approaches have considered entropy-based metrics for clustering boat activity into three broad categories: random walk, following, and chasing. Here, we comprehensively assess the accuracy of neural network-based approaches as alternatives to entropy-based clustering. We train four neural network models and compare them to shallow learning using synthetic data. We also investigate the accuracy of models as time steps increase and with and without rotated data. To improve test-time robustness, we normalize trajectories and perform rotation-based data augmentation. Our results show that deep networks can achieve a test-set accuracy of up to 100\% on a full trajectory, with graceful degradation as the number of time steps decreases, outperforming entropy-based clustering.
\end{abstract}

\keywords{Maritime Threat Detection \and Trajectory Classification \and Neural networks \and Threat Detection}

\section{Introduction}

Illegal maritime activities have increased over the years, and the demand for early maritime threat artificial intelligence detection systems has increased. \cite{desai2021, martinez-zarzoso2010} Most boats are equipped with a radar that tracks the coordinates of surrounding boats. \cite{fowdur2021} An automated approach that assesses potential threats could then be used based on the boat's trajectory. A straightforward option is to classify trajectory type and threat level based on predefined labels such as chasing, following, and non-threatening as used by Chen et al. \cite{chen2022}. Ideally, such classification systems could detect the trajectory type based on a small observation window and operate with minimal overhead. Training such classifiers necessitates the use of maritime trajectory data exhibiting these labels. However, existing radar and satellite data sets are often incomplete in nature and do not always contain behaviors exhibited by malicious actors \cite{chen2022}.

An alternative approach could be to use synthetic data that models expected trends in both innocuous and malicious boat activity in order to train classification models. Kassab et al. discuss the benefits of using synthetic datasets for general maritime purposes, especially in the case of marine threat detection, as models are more robust and more applicable to the real world compared to models that train on sparse and inaccurate datasets \cite{kassab2020}. In addition, synthetic datasets are often limited by the rules from which they are created and may not accurately represent real-life \cite{Goyal24}. In order to better generalize these synthetic data sets and, therefore, make the detection medium more accurate, data augmentation, such as rotation, can be applied to the data, leading to a more diverse and applicable dataset \cite{app}.

Several approaches could be used to classify boat activity. In particular, the use of artificial intelligence to accurately automate the detection of these malicious actors could save precious time and improve maritime defense \cite{ADF2024, AIFSG2024}. In the present work, we explore the use of neural network-based classifiers on synthetic data representing boat trajectories. We evaluate the performance of different neural network architectures in terms of classification accuracy. We explore how the accuracy changes as a function of trajectory length, and we compare to previously proposed classical approaches to classification. We find that neural networks improve classification accuracy, especially for short trajectories, highlighting the potential for deep learning in maritime threat detection.

\subsection{Related Work}

Several works have explored approaches for identifying malicious trajectories. Most automatic identification system (AIS) data has been used to monitor criminal activities such as drug smuggling and piracy in the military domain \cite{jmse10010112}.

Obradovic et al. proposed to detect anomalies from regular boat trajectories using AIS data \cite{obradovic2014machine}. Unfortunately, the data are often misrepresented and malicious actors can easily transmit false AIS information. Moreover,  data detailing malicious activities is lacking. Also, the proposed method fails to explain the reasoning behind the detection or what behavior a certain actor exhibited that led to its detection. The lack of data also means that the model that is trained on the data is unable to generalize, meaning that when given new data, it is less likely to be able to predict it accurately. Futhermore, there is a lack of early detection, meaning that even if insights are discovered and explained, it is often too late for users to act on them.

Chen et al. proposed entropy-based clustering algorithms to detect synthetic trajectories early \cite{chen2022}. These clustering algorithms tend to have high accuracies and apply to a wide range of trajectory lengths. These models will need to be directly compared to the proposed neural network approach to determine the state-of-the-art model types on synthetic trajectory data.

Strauch et al. demonstrated that the onboard sensor information, combined with intelligence from external sources, proved valuable for early piracy threat detection \cite{strauch2021overhead}. This research aims to create a model and identify the best model type for training on synthetic coordinate data to apply on information returned by these sensors.

\subsection{Our Contribution}

Our present work is the first to evaluate the use of a variety of deep neural network types on synthetic data for maritime threat detection. Our models achieve the accuracy of 100\% on a full trajectory while also performing as well as 87\% with only 1/12th of the full trajectory. The present work is also the first to evaluate the use of convolutional neural networks and Dense models in the field of trajectory prediction, where Recurrent Neural Networks would likely be preferred.

In addition, the presently reported is a dramatic improvement on the previous entropy-based clustering algorithms proposed by Chen et al. We have significantly improved on the previous entropy-based clustering algorithms proposed by Chen et al on a variety of benchmarks. On full trajectories, our models achieved 100\% accuracy while the clustering algorithms could, at best, only achieve 97\%. On incomplete trajectories, the clustering algorithms achieved relatively low accuracies around (60 - 70\%) while the proposed neural networks performed high accuracies around 80 to 100\%.

Finally, our research discusses a lesser-known behavior with RNN models such as the LSTM or GRU and a similar behavior among other models where rotating the data makes the models perform poorly. We explore the possible mechanisms behind the behavior and the implications of this behavior in maritime threat detection systems and a variety of fields.

\section{Methods}

We consider the setting consisting of two boats in a shared coordinate system. Boat $U$ is considered to be a malicious actor, and Boat $V$ is considered to be the target of Boat $U$. Following \cite{chen2022}, we use a simulation framework in which boat $U$ moves according to one of three prescribed strategies: random walk, chasing, or following. We assume that boat $U$ is able to measure the coordinates of the other boat at any given time point and take action based on this information as malicious actors also have access to the same information. We further assume that each boat moves according to one of three prescribed strategies: random walk, chasing, and following. Using the trajectory information, the goal is for Boat $V$ to classify the trajectory type of Boat $U$.

\subsection{Trajectory Simulation}

In order to simulate aggression in malicious trajectories, following Chen et al. \cite{chen2022}, Boat $U$ uses the Bresenham algorithm to track and pursue boat $V$ \cite{koopman1987bresenham}. The Bresenham algorithm is used to draw the approximation of the line that represents the shortest path possible between the two boats based on predictions of the future position of a target boat \cite{koopman1987bresenham}. Boat $U$ moves according to one of three prescribed strategies:

\textbf{Random Walk:} Random Walk is considered to be the trajectory of an innocent actor in the generated synthetic data. Every 5 time steps, the simulation randomly selects a cardinal direction for boat $U$ to move in after waiting a random amount of time. 

\textbf{Chasing:} Boat $U$ moves towards boat $V$ using the line created by the Bresenham line drawing algorithm. The speed is randomized in every instance. 

\textbf{Following:} Boat $U$ ensures that it is always behind boat $V$ while randomly choosing one of three movement types: zigzag, chasing, and random-walk at random intervals. 

The coordinates of the data are recorded from a C++ simulation and are then exported to a .csv file to be loaded into a separate Python workflow for analysis. 

Each trajectory consists of the coordinate data of Boat $U$ when chasing boat $V$. The goal of Boat $U$, while undergoing a chasing or following trajectory, is to catch $V$. Boat $U$ will always start on (2048, 2048) and will have 6000 time steps to catch the boat $V$. Boat $V$ will start on a random point within the boundaries (0,0) and (4096, 4096) and move one unit to the east every 5 time steps.

The simulation ends when boat $V$ is caught, the time limit of 6000 timesteps is reached, or the malicious boat travels out of bounds. In total, 100 trajectories of each trajectory type, random walk, chasing, and following, are generated, thus resulting in a data set of 300 trajectories.

\begin{figure}[h!]
    \centering
    \includegraphics[width=0.5\linewidth]{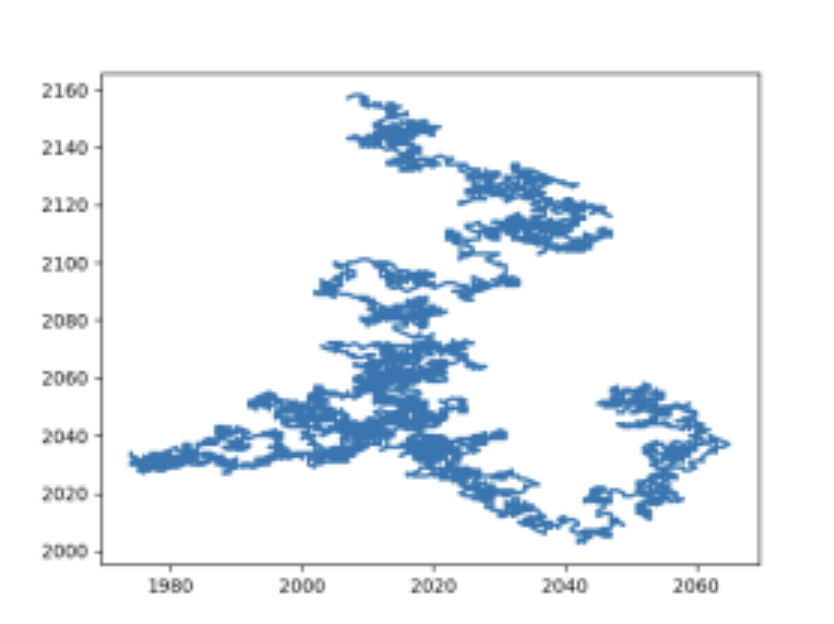}
    \caption{A Complete Random Walk Trajectory}
    \label{fig:random}
\end{figure}

\begin{figure}[h!]
    \centering
    \includegraphics[width=0.5\linewidth]{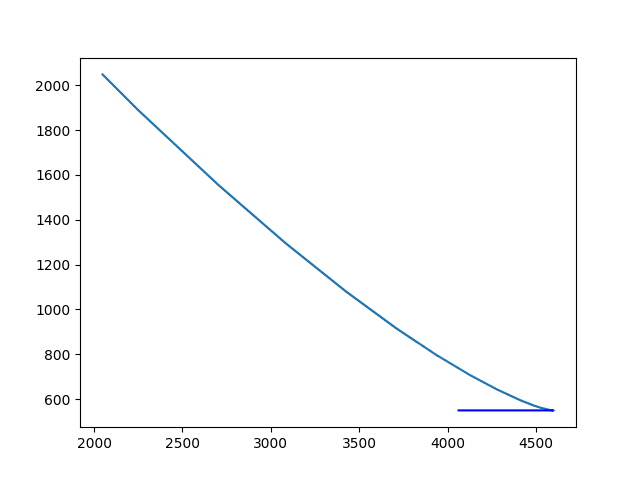}
    \caption{A Complete Chasing Trajectory}
    \label{fig:chasing}
\end{figure}

\begin{figure}[h!]
    \centering
    \includegraphics[width=0.5\linewidth]{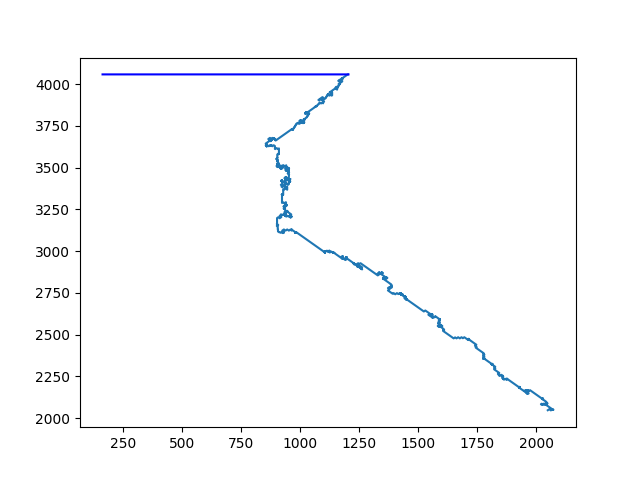}
    \caption{A Complete Following Trajectory}
    \label{fig:following}
\end{figure}

Figures~\ref{fig:random}, \ref{fig:chasing}, ~\ref{fig:following} show examples of random walk, chasing, and following trajectories, respectively, where Boat $V$ is shown in the darker blue shade. The random walk trajectory is not shown with Boat $V$ as it never interacts with Boat $V$ and its shape will not be properly visible if also displaying Boat $V$.

\subsection{Data Preprocessing}

The data are preprocessed before the models train on it.
First, 100 instances of each trajectory type are put into one dataset. Its corresponding labels are also put in a separate dataset. Zero is assigned for Random Walk, 1 is assigned for Chasing, and 2 is assigned for a Following trajectory. 

Second, the data is loaded into a Numpy array and randomized. Fifteen trajectories of each trajectory type are randomly chosen and put in a test set used for later evaluation. The datasets are then randomized, and exported to CSV for later evaluation of the clustering algorithms of \cite{chen2022}. 

In addition, a rotated version of a train and test set is exported in a separate CSV file for evluation using procedure described below.

\textbf{Rotation:} Each trajectory is rotated by a distinct angle by the matrix shown below in equation \ref{eq:rotation}. All train and validation data is rotated randomly when the model is being evaluated on a rotation dataset.  

\begin{equation}
\label{eq:rotation}
\begin{bmatrix}
\cos \theta & -\sin \theta \\
\sin \theta & \cos \theta
\end{bmatrix}
\end{equation}\\

\subsection{Model Types}

We consider four neural network-based architectures for a trajectory classification task. Each network takes the trajectory coordinates of Boat $U$ up to $T$ timesteps as input and outputs a predicted label.

\subsubsection{Dense}

A dense model, also known as a fully connected network, completely maps the input to the output. The model consists of neurons connected by weights and tuned as it trains to provide more accurate detections. Dense layers are also used to map outputs from the LSTM, GRU, and CONV1D to predictions \cite{wang2017time}.

\subsubsection{LSTM}

An LSTM or Long Short Term Memory is a type of recurrent neural network (RNN) that is designed to overcome the issues of traditional RNNs particularly in the case of the vanishing/exploding gradient problem. The vanishing/exploding gradient problem occurs when the gradient of the model exponentially decreases or increases during backpropagation, making the model unviable for long-term dependencies. The LSTM mitigates this problem by using its long-term memory while still containing short-term memory \cite{cho2014}.

\subsubsection{GRU}

Similar to the LSTM, a Gated Recurrent Unit or GRU is a type of RNN that solves the vanishing/exploding gradient problem. In addition, the GRU model architecture specializes in sequential data and is comparable to the LSTM in accuracy (Junyoung et al.). However, the GRU architecture is not a complete model; rather, is an individual trainable unit comparable to the tanh or sigmoid functions, but it still can be used as a complete model. The GRU architecture makes use of two simple gates: the update gate and the reset gate \cite{hochreiter1997}.

\subsubsection{Conv1D}

Unlike the LSTM or GRU model architectures, Conv1D models are convolutional models that apply a kernel of trained values to the data. The kernel size is 2, and the values change as the model is trained. The values that come from the data are then flattened and used by Dense layers to create a prediction \cite{zheng2014time}.

\subsection{Model Training}

\textbf{K-fold cross-validation}: Cross-validation of K-fold is a technique used to evaluate and improve the performance and generalizability of a machine learning model. It involves partitioning the training and validation dataset into subsets, called folds. In this research, five folds are utilized. One fold is chosen to be the validation set used to evaluate the model. The rest of the data is used as the training set. A new model is trained on the corresponding training set and its final accuracy on the validation set is recorded. The process repeats until all of the data has been a validation set and five distinct models have been trained. The model with the highest validation accuracy is chosen to be evaluated on the test set.

\textbf{Timesteps}: In each of the experiments the model is trained and evaluated at different times of the trajectories, called timesteps. Each trajectory has 6000 timesteps, and the model is trained on every 500 time steps from 500 to 6000.

\begin{figure}[h!]
    \centering
    \includegraphics[width=0.75\linewidth]{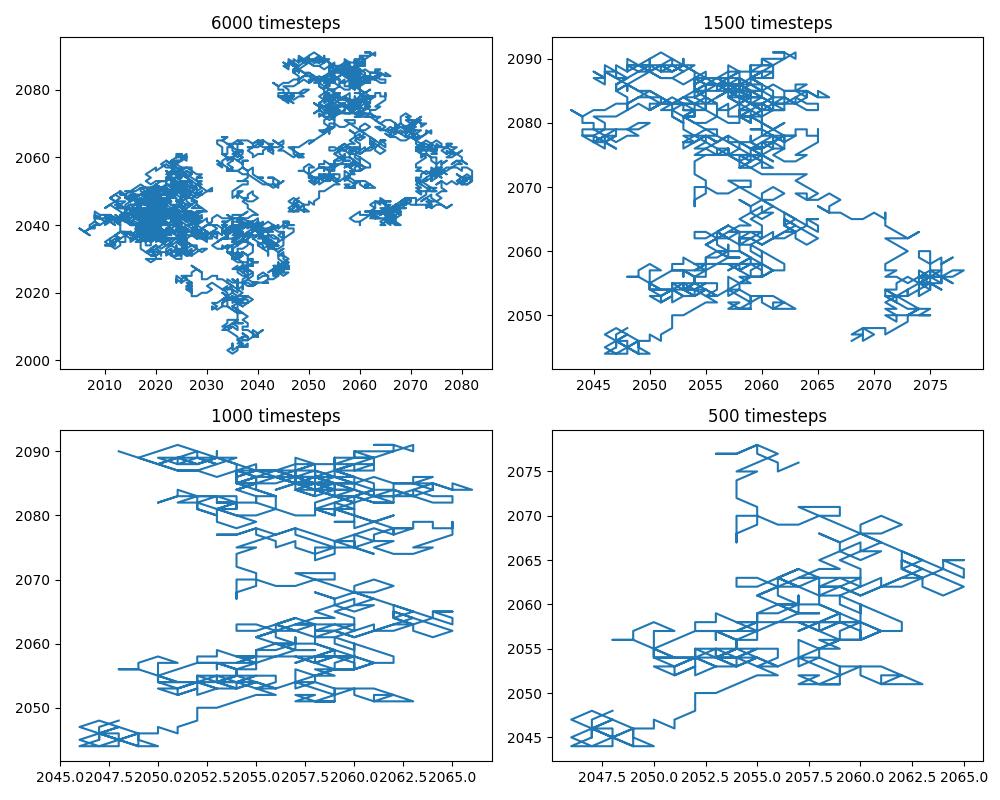}
    \caption{A Visualization of a Random Walk Trajectory at different timesteps}
    \label{fig:random_walk_collage}
\end{figure}

\begin{figure}[h!]
    \centering
    \includegraphics[width=0.75\linewidth]{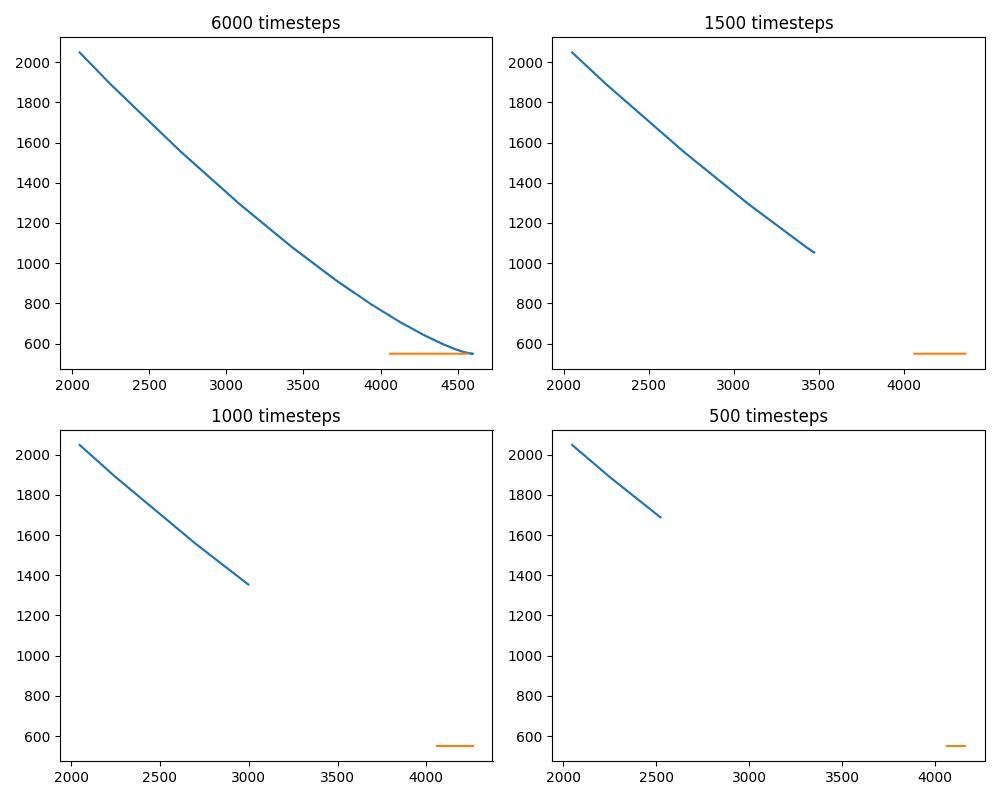}
        \caption{A Visualization of a Chasing Trajectory at different timesteps}
    \label{fig:Chase_walk_timestep_collage}
\end{figure}

\begin{figure}[h!]
    \centering
    \includegraphics[width=0.75\linewidth]{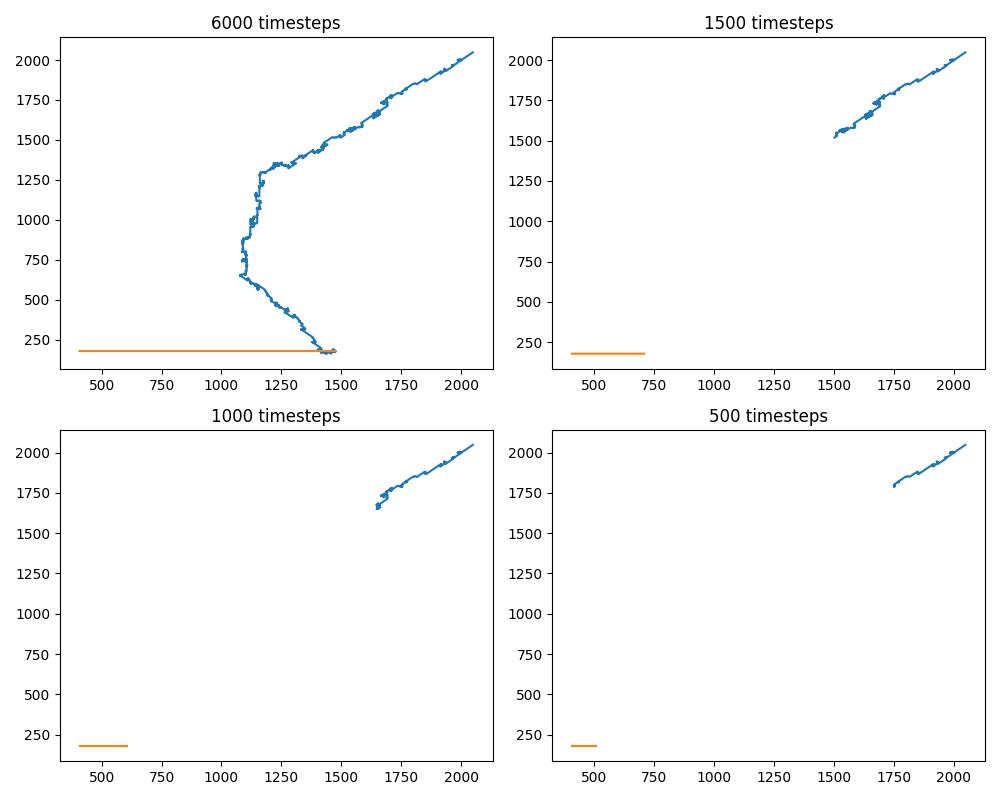}
    \caption{A Visualization of a Following Trajectory at different timesteps}
    \label{fig:follow_walk_timestep_collage}
\end{figure}

Figures~\ref{fig:random_walk_collage}, \ref{fig:Chase_walk_timestep_collage}, and \ref{fig:follow_walk_timestep_collage} show a random walk, chasing, and following trajectories, respectively, at different timesteps. The other line visible on the chasing and following trajectory represents boat $V$.

\textbf{Main model training}: A training set for each of the timestep groups is created. Then, the training dataset is fed into the k-fold cross-validation algorithm. Then the model is trained on the returned train set and evaluated on the returned validation set. This repeats for every fold until the best model is returned. The best model is then evaluated on the test set and the accuracy is recorded. This repeats for every time step group and with and without rotated data.

\subsubsection{Entropy-based Clustering Model Training}

A comparison experiment has been conducted to evaluate the effectiveness of the proposed neural networks compared to the state-of-the-art clustering algorithms.

First, the train set and rotated train set are exported into CSV to generate new entropies using a history buffer of size $b$. The entropies are generated after estimating the probabilities. 

For the supervised methods, we evaluate the average of the buffer entropy data points to get the entropy value for each trajectory. We then generate the thresholds for the clustering algorithm and the supervised algorithm for a full trajectory for both the rotated and non-rotated data. 

For the KMeans algorithms, $Q$ centroids are added, per the KMeans algorithm, and the final thresholds, representing the mean of the dataset, are returned for both the non-rotated and rotated dataset.

Finally, the clustering algorithms are tested on every timestep group with its corresponding thresholds, and the accuracies are recorded. In order to test the clustering and unsupervised algorithms with rotation, the same steps are repeated above but with the rotated dataset.

\section{Results}

\subsection{Main Results}

The accuracies of the models on a separate test set are reported in this subsection here.

First, the accuracies for the Dense model trained on a rotated and non-rotated set on the test set are recorded as shown in Figures \ref{fig:nraccuracy} and \ref{fig:raccuracy}. The Dense model achieves a constant increase and consistently reaches high accuracies above 90\% for timesteps as low as 1000.

Second, the accuracies on the test set for the recurrent neural networks, GRU and LSTM, are shown in Figures \ref{fig:nraccuracy} and \ref{fig:raccuracy}. Overall, the best RNN is the LSTM, which has higher accuracies across the board. However, both models struggle when the data is rotated, with considerably lower accuracies, barely reaching the 70\% range. On the other hand, LSTM performs extremely well when training on a non-rotated se,t reaching 100\% as soon as 3000 timesteps. In fact, the LSTM performs the best out of all of the models on the non-rotated set.

Finally, the Conv1D accuracies on the test set are displayed in Figures \ref{fig:nraccuracy} and \ref{fig:raccuracy}. The accuracies of the model are very high compared to the LSTM and GRU. However, overall, the Conv1D model's accuracies are not as high as the accuracies returned by the Dense model. Granted, the Conv1D model has slightly higher accuracies at earlier timesteps, like 1000 and 1500 but overall, the Dense model still returns higher accuracies.

\subsection{Clustering Algorithm Results}

First, using the training set, the thresholds shown in Table \ref{tab:Thresholds} are generated for the clustering algorithm.

\begin{table}[h!]
\centering
\begin{tabular}{|c|c|c|c|c|}
\hline
& \textbf{Supervised} & \textbf{Supervised} & \textbf{K-means} & \textbf{K-means} \\
& \textbf{Weighted} & \textbf{Voronoi} & \textbf{Weighted} & \textbf{Voronoi} \\ \hline
$T_1$ & 0 & 0 & 0 & 0 \\ \hline
$T_2$ & 1.414 & 1.797 & 1.421 & 1.810 \\ \hline
$T_3$ & 3.057 & 2.819 & 3.027 & 2.816 \\ \hline
\end{tabular}
\caption{The Thresholds Generated from the Train Set}
\label{tab:Thresholds}
\end{table}

\begin{figure}[h]
    \centering
    \includegraphics[width=0.5\textwidth]{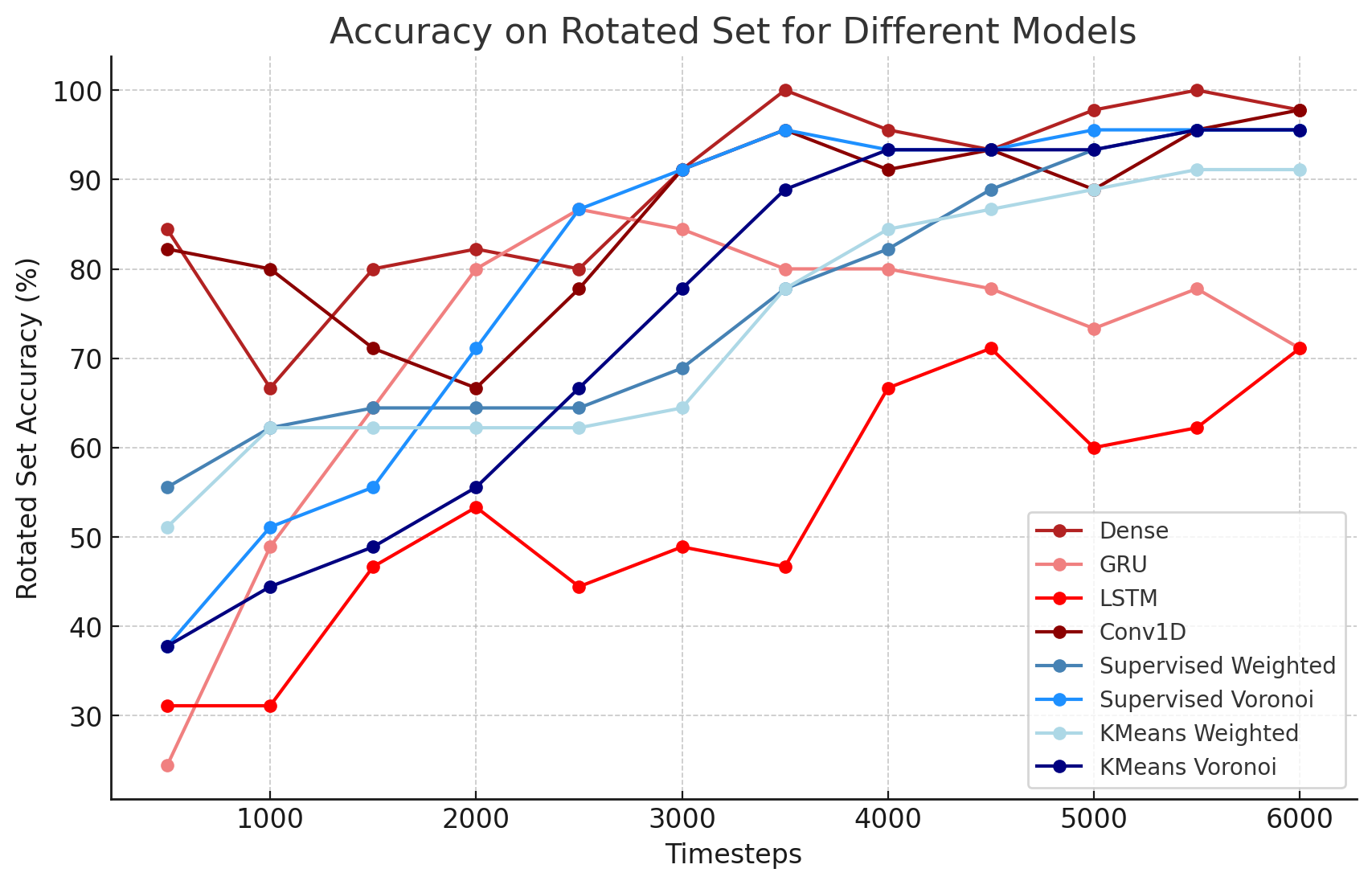}
    \caption{Accuracy on Rotated Test Set for Different Models}
    \label{fig:raccuracy}
\end{figure}

\begin{figure}[h]
    \centering
    \includegraphics[width=0.5\textwidth]{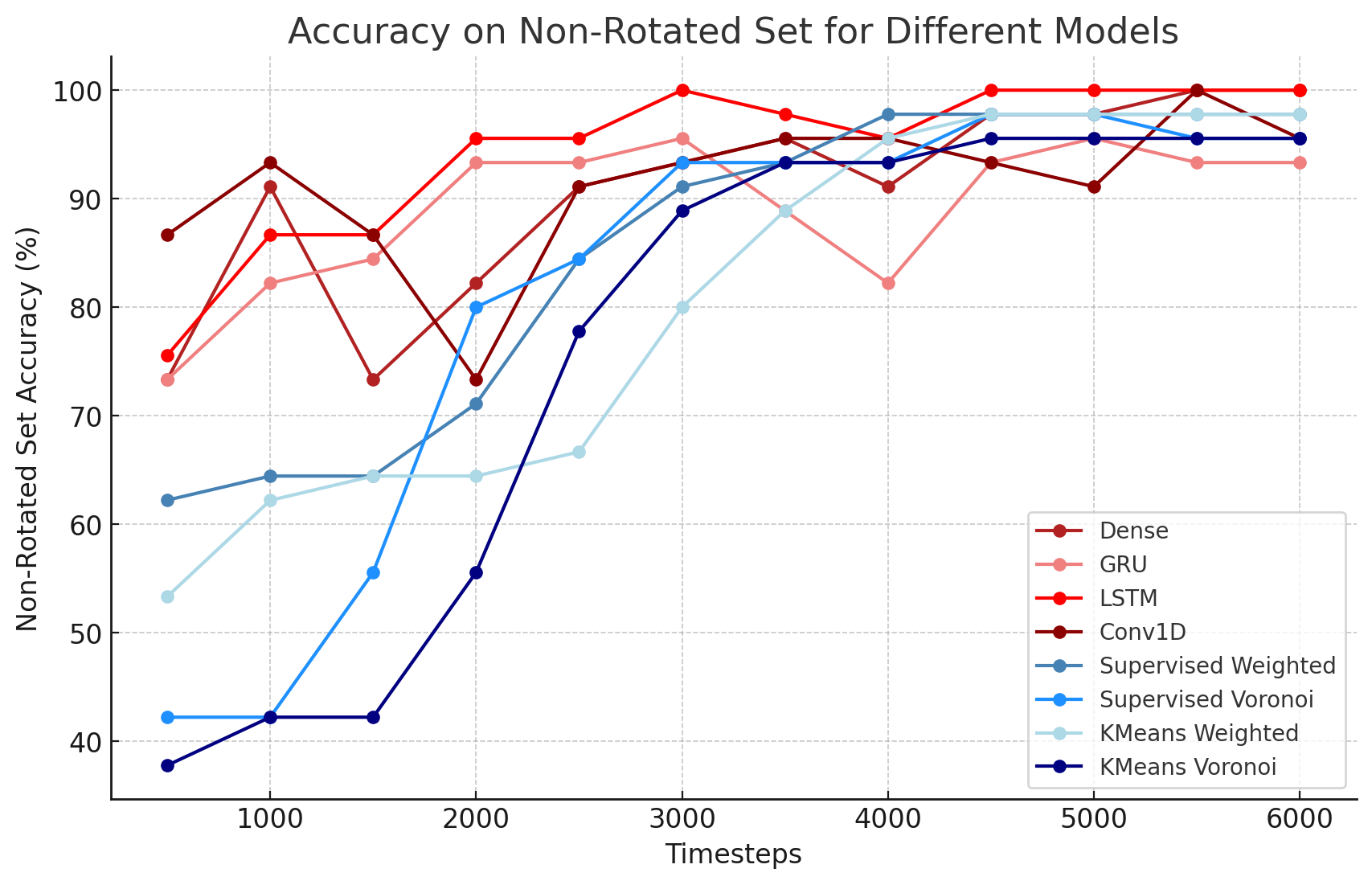}
    \caption{Accuracy on Non Rotated Test Set for Different Models}
    \label{fig:nraccuracy}
\end{figure}

Then, the thresholds are used to evaluate the test set and return accuracies as shown by the red shaded lines in Figures \ref{fig:nraccuracy} and \ref{fig:raccuracy}.

A general trend with the clustering algorithms is that they do not perform well when timesteps are low. However, they always see a consistent increase in accuracy as timesteps increase, unlike the proposed neural networks, which sometimes drop in accuracy.

We run the thresholds across the same timestep groups as the proposed neural networks are evaluated for both the rotated and non-rotated sets.

The accuracies for the Supervised Weighted clustering algorithm, as shown by the line labeled Supervised Weighted in Figures \ref{fig:nraccuracy} and \ref{fig:raccuracy}, are higher when being evaluated on a non-rotated test set. However, both the non-rotated and rotated test set accuracies eventually reach above 95\%. However, compared to the Conv1D or the Dense models, the accuracies are comparably lower at almost every timestep. Even though the accuracies on the non-rotated test set are still high, compared to the LSTM, they are considerably lower.

The Supervised Vonoroi accuracies, as shown by the line labeled Supervised Vonoroi in Figures \ref{fig:nraccuracy} and \ref{fig:raccuracy}, are slightly lower than the accuracies of the Supervised Weighted model across the board. 

The accuracies for the KMeans Clustering Algorithms are shown by the lines labeled with KMeans in as shown by the line labeled SWeighted in Figures \ref{fig:nraccuracy} and \ref{fig:raccuracy}.

Overall, the KMeans Clustering algorithms performed much worse on the rotated set than the other Supervised Clustering algorithms and the proposed neural networks, especially on lower timestep groups. The KMeans Voronoi algorithm performs a little better compared to the KMeans Weighted algorithm on the rotated set.

On the non-rotated set, the KMeans Vonoroi algorithm performed slightly better than the KMeans Weighted algorithm. However, compared to other clustering algorithms and the proposed neural networks, the accuracies are considerably lower, especially at lower timestep groups.

Overall, on almost all timesteps, the proposed neural networks perform better than the clustering algorithms.

\section{Discussion}


The Dense models were the most robust models on average, with high accuracies across the board. At lower time steps, the models were comparably extremely accurate, reaching accuracies as high as 80\% and 91\% on the rotated and non-rotated sets, respectively. At higher time steps, the Dense models consistently had high accuracies from 90\% to 100\%.

The Conv1D models performed well on both the non-rotated and rotated datasets but did not perform as well as the Dense models overall. The Conv1D model was initially better and more consistent at lower time steps than the Dense model, with a max accuracy of 93\% compared to the max accuracy of 91\%.

The LSTM was the most consistent and accurate model on non-rotated datasets. At higher time steps, the LSTM managed almost always to achieve an accuracy of 100\%. At lower time steps, the LSTM models outperformed the Dense models with more consistent and higher accuracies on average. Overall, the LSTM showed the most consistent increase in accuracy as time steps increased and had the highest accuracies on the non-rotated set. However, the LSTM performed subpar on the rotated data sets with only a slight increase as time steps increased.

The GRU, an RNN similar to the LSTM, was less consistent and accurate as time steps increased compared to the LSTM but still achieved high levels of accuracy on the non-rotated sets. Also like the LSTM, the GRU performed relatively worse on the rotated datasets but still performed better than the LSTM overall.

The clustering algorithms return substantially lower accuracies on both the rotated and non-rotated sets compared to the models. The highest accuracies that they return are on the non-rotated sets where the LSTM, Dense, and Conv1D perform considerably higher. However, the clustering algorithms see a more consistent increase in accuracy as timesteps increase.

\subsection{Considerations}

RNN models, like the GRU and LSTM, usually perform the best when dealing with sequential data, such as the data used in this research. However, rotating the data before the model is trained on them disrupts the model and results in lower accuracies. This behavior is due to two possible reasons. 

First, RNN models like the LSTM and GRU do not explictly consider the distance between data points or spatial data. Rotating the data disrupts the spatial properties of the data which prevents the models, which do not have the means to adapt to the spatial ambiguity, from generalizing well and achieving high accuracies. In other words, the LSTM and GRU models assume a constant distance between points to achieve high accuracies but rotating the data changes the distance between points and makes it harder for the models to adapt.

Second, the LSTM and GRU expect certain data features to be in a certain format or location. For example, all non-rotated trajectories start at the coordinate (2048, 2048). However, rotating the data makes these locations ambiguous, which makes it harder for the model to generalize and learn \cite{shiri2023}.

Overall, our research demonstrates that RNN models like the LSTM and GRU are less suitable for trajectory prediction tasks where spatial orientation, speed in the real world, and starting locations are ambiguous. This research also confirms that the LSTM and GRU perform worse with prepossessing techniques like rotation and emphasizes the need for the development of a sequential model with a mechanism to learn regardless of ambiguity in spatial data and unknown starting locations.

This behavior is not limited to the LSTM or GRU, as other models experienced a smaller but similar drop in accuracy from the rotated and non-rotated sets. This behavior most likely occurred because the other models are better equipped when dealing with varying spatial orientations but still struggle with the ambiguity of attributes of the data, like the starting location.

When deciding on-on the best model, the computational complexity of the models must also be considered, as on-board computers must be able to apply multiple instances of the model to evaluate the threat from multiple actors. The Dense model, the most accurate model on rotated datasets, is the most computationally inexpensive making it an even better candidate for this application. In addition, the predictions from the Dense models are rapid and can be done in any number of time steps while maintaining relatively high accuracy.

\subsection{Applications}

As mentioned earlier, our work's results can be used in the domain of general maritime threat prediction systems. Since the model explains the reason behind why an actor is declared malicious, i.e., because they were doing a chasing or following trajectory, modern detection systems have a better understanding and comprehension of the subject.

This research emphasizes the use of the Conv1D and Dense model for trajectory classification instead of the traditional approach of using Recurrent Neural Networks (RNN) like the LSTM and GRU. In other words, instead of only using RNNs, the Conv1D and Dense models can also be considered for use in cases where trajectory classification is needed such as malicious actor detection in crowded environments and autonomous driving applications.

Finally, the models proposed in this research can almost be considered for any application with trajectory classification inside and outside maritime detection systems, like whether the trajectory of an object is threatening in autonomous driving scenes or whether an actor's trajectory is malicious in security camera footage.

\subsection{Limitations}

Neural Networks are great when it comes to classifying trajectories with a fixed amount of data. However, when looking at the data in the real world, where there are an infinite amount of lengths that a trajectory can have, neural networks can not be applied directly since they require to be applied on trajectories with the exact size as trajectories that they are trained on, which is not realistic.

In the comparison study, the entropies and thresholds generated for the clustering algorithms were not generated on the exact training split but were still generated on the same dataset.

\subsection{Future Work}

In the future, models should be applied and evaluated on real modern detection systems. The models should also be evaluated on real-life datasets to emphasize the applicability of synthetically trained models in real-life scenarios. More types of models, such as the Transformer or Conv2D framework, should be tested on the application. Finally, more data augmentation techniques like scaling and translation, aside from rotation, should be tested.

\section{Conclusion}
Illegal maritime activities are on the rise, and there is a need for automated detection systems. Unfortunately, existing datasets are sparse and inaccurate. In addition, precious approaches fail to detect malicious actors early enough. Therefore, this research proposes to create and rotate synthetic data to train four distinct models at a variety of proportions of the full trajectory.

The previous state-of-the-art method proposed in Chen et al. and classical clustering have been compared to the proposed neural networks. Our proposed approach performs substantially better, reaching accuracies of 82\% at lower timesteps and even 100\% at higher timestep groups. These results suggest that neural networks, when trained with rotated synthetic data, are promising for improving maritime threat detection systems and can be generalized to other domains where timely anomaly detection is critical.

\section*{References}
\bibliographystyle{plain}

\end{document}